\documentclass[11pt]{article}

\usepackage{amsmath}
\usepackage{amsfonts}
\usepackage{epsfig}
\usepackage{subfigure}
\usepackage{fancybox}
\usepackage{my-apa-uiuc}

\begin{document}
\sloppy
\begin{titlepage}
\hspace{0.08in}
\begin{minipage}{\textwidth}
\vspace*{2.3in}
\begin{center}
{\bf Sub-Structural Niching in\\
 Estimation of Distribution Algorithms}\\
\addvspace{0.5in}
{\bf Kumara Sastry}\\
{\bf Hussein A. Abbass}\\
{\bf David E. Goldberg}\\
{\bf D.D. Johnson}\\
\addvspace{0.3in}
IlliGAL Report No. 2005003 \\
February, 2005\\
\vspace*{-0.1in}
\vspace*{3in}
Illinois Genetic Algorithms Laboratory \\
University of Illinois at Urbana-Champaign \\
117 Transportation Building \\
104 S. Mathews Avenue
Urbana, IL 61801 \\
Office: (217) 333-2346\\
Fax: (217) 244-5705 \\
\end{center}
\end{minipage}
\end{titlepage}

\title{Sub-structural Niching in Estimation of Distribution Algorithms}
\author{Kumara Sastry$^1$, Hussein A. Abbass$^2$, David
  E. Goldberg$^1$, D.D. Johnson$^3$\\
~\\
$^1$Illinois Genetic Algorithms Laboratory (IlliGAL),\\ 
Department of General Engineering,\\
 University of Illinois at Urbana-Champaign\\
~\\
$^2$Artificial Life and Adaptive Robotics Laboratory,\\
 School of Information Technology and Electrical Engineering,\\
University of New South Wales, Australian Defense Force Academy\\
~\\
$^3$Department of Materials Sciences and Engineering,\\
University of Illinois at Urbana-Champaign\\
~\\
{\tt ksastry@uiuc.edu, h.abbass@adfa.edu.au, deg@uiuc.edu,
         duanej@uiuc.edu}
}
\date{}
\maketitle
\begin{abstract}
  We propose a sub-structural niching method that fully
  exploits the problem decomposition capability of linkage-learning
  methods such as the estimation of distribution algorithms and
  concentrate on maintaining diversity at the sub-structural level.
  The proposed method consists of three key components: (1) Problem
  decomposition and sub-structure identification, (2) sub-structure
  fitness estimation, and (3) sub-structural niche preservation. The
  sub-structural niching method is compared to restricted tournament
  selection (RTS)---a niching method used in hierarchical Bayesian
  optimization algorithm---with special emphasis on sustained
  preservation of multiple global solutions of a class of
  boundedly-difficult, additively-separable multimodal problems. The
  results show that sub-structural niching successfully maintains
  multiple global optima over large number of generations and does so
  with significantly less population than RTS. Additionally,
  the market share of each of the niche is much closer to the expected
  level in sub-structural niching when compared to RTS.
\end{abstract}

\section{Introduction}
One of the daunting challenges in the field of genetic and
evolutionary computation is the systematic and principled design of
scalable genetic algorithms (GAs) and significant progress has been
made along these lines. A design decomposition theory has been
proposed and several {\em competent} GAs---GAs that solve hard
problems quickly, reliably, and accurately---have been developed
\cite{Goldberg:1999:Race}. One such class of competent GAs is the
estimation of distribution algorithms (EDAs)
\cite{Pelikan:2002:EDAsurvey,Larranaga:2002:EDAbook}. EDAs replace the
traditional variation operators of GAs with probabilistic model
building of promising solutions that identifies key sub-structures (or
building blocks) of the underlying search problem, and sampling the
model to generate new candidate solutions. EDAs have successfully
solved problems of bounded difficulty at a single level or at multiple
hierarchical levels requiring only polynomial (oftentimes
sub-quadratic) number of function evaluations
\cite{Pelikan:2000:BOA,Pelikan:2001:hBOA,Pelikan:2003:BOAscalability}.\par

One of the important components required by EDAs for successfully
solving multimodal, hierarchical, dynamic, and multiobjective
optimization problems is an efficient niching method. The niching
mechanism is required to stably maintain a diverse population
throughout the search, thereby allowing EDAs to (1) identify multiple
optima reliably when solving multimodal and multiobjective problems,
(2) identify the global optimum by deciding successfully between
sub-structures when all the hierarchical interactions are revealed,
and (3) rapidly identify global solutions as and when changes occur in
non-stationary problems. Such a niching method not only needs
to adaptively identify and conform to all the niches and
niche-distance distributions, but also need to maintain
them effectively over the duration of the search.\par

Traditional niching methods usually maintain diversity at the level of
individuals and are not often adaptive to the niche size and
distribution. Additionally, they also do not exploit the underlying
working mechanism of EDAs and other linkage-learning algorithms. That
is, the traditional nichers often use distance information based on
the entire individual and do not often directly respect or exploit
problem decomposition. Therefore, in this paper we propose a niching
method that respects problem decomposition, and utilizes the
sub-structure identification capability of EDAs, and maintains
diversity at sub-structure level in a stable manner. Such a niching
method is not only advantageous for maintaining multiple niches, but
also effective for hierarchical \cite{Pelikan:2001:hBOA}, and dynamic
\cite{Branke:2001:dynamic,Abbass:2004:dynamic} problem optimizations
where sub-structure niche preservation is what actually required.\par

The proposed method consists of three components: (1) Sub-structure
identification, where we use the probabilistic model built by EDAs,
specifically, extended compact GA \cite{Harik:1999:eCGA}, (2)
sub-structure fitness estimation, where we use the fitness-estimation
procedure proposed by Sastry {\em et al\/}
\cite{Sastry:2004:eCGAinheritance}, and (3) sub-structure niche
preservation, where different mechanisms can be envisioned, and
suitability of each is based on the purpose and objective of
niching. The key idea of the sub-structure niche preservation
mechanism is to preserve highly-fit sub-structures in desired
proportions in the population in a stable manner over the duration of
the search.\par

The sub-structural niching mechanism is compared with restricted
tournament selection \cite{Harik:1995:RTS}---a nicher used in
hierarchical Bayesian optimization algorithm (hBOA)---on a class of
boundedly-difficult additively-separable multimodal
problems. Specifically, we compare (1) the stability of maintaining
multiple niches over a large number of generations, (2) the capability
of allocating market share to different niches at the desired level,
and (3) the population size required to consistently maintain all the
global optima.\par

The paper is organized as follows. The next section provides a brief
introduction to the extended compact genetic algorithm, followed by a
detailed description of the proposed sub-structural niching mechanism.
The performance of the proposed method is compared to that of RTS in
section~\ref{sec:results}, followed by key conclusions of the paper.

\section{Extended Compact Genetic Algorithm}
Extended compact genetic algorithm (eCGA) \cite{Harik:1999:eCGA} is an
EDA that replaces traditional variation operators of genetic and
evolutionary algorithms by building a probabilistic model of promising
solutions and sampling the model to generate new candidate solutions.
The typical steps of eCGA can be outlined as follows:
\begin{enumerate}
\item {\em Initialization:\/} The population is usually initialized with
  random individuals. However, other initialization procedures can also
  be used in a straightforward manner.
\item {\em Evaluation:\/} The fitness or the quality-measure of the
  individuals are computed.
\item {\em Selection:\/} Like traditional genetic algorithms, EDAs
  are selectionist schemes, because only a subset of better
  individuals is permitted to influence the subsequent generation of
  candidate solutions. Different selection schemes used elsewhere in
  genetic and evolutionary algorithms---tournament selection,
  truncation selection, proportionate selection, etc.---may be adopted
  for this purpose, but a key idea is that a
  ``survival-of-the-fittest'' mechanism is used to {\em bias\/} the
  generation of new individuals.
\item {\em Probabilistic model estimation:\/} Unlike traditional
  GAs, however, EDAs assume a particular probabilistic model of the
  data, or a {\em class\/} of allowable models. A {\em class-selection
    metric\/} and a {\em class-search mechanism\/} is used to search
  for an optimum probabilistic model that represents the selected
  individuals.\par
 
  \noindent
  {\bf Model representation:} The probability distribution used
  in eCGA is a class of probability models known as marginal product
  models (MPMs). MPMs partition genes into mutually independent
  groups and specifies marginal probabilities for each linkage group.\par

  \noindent
  {\bf Class-Selection metric:} To distinguish between better model
  instances from worse ones, eCGA uses a minimum description length
  (MDL) metric \cite{Rissanen:1978:MDL}. The key concept behind MDL
  models is that all things being equal, simpler models are better
  than more complex ones. The MDL metric used in eCGA is a sum of two
  components:
  \begin{itemize} 
  \item {\bf Model complexity} which quantifies the model representation
    size in terms of number of bits required to store all the marginal
    probabilities:
    \begin{equation}
      \label{eqn:cm}C_m = \log_2(n)\sum_{i = 1}^{m}\left(2^{k_{i}} - 1\right).
    \end{equation}
    where $n$ is the population size, $m$ is the number of linkage
    groups, $k_i$ is the size of the $i^{\mathrm{th}}$ group.
  \item {\bf Compressed population complexity}, which quantifies the data
    compression in terms of the entropy of the marginal distribution
    over all partitions.
    \begin{equation}
      C_p = n\sum_{i = 1}^{m}\sum_{j = 1}^{2^{k_{i}}} -p_{ij}\log_2\left(p_{ij}\right),
    \end{equation}
    where $p_{ij}$ is the frequency of the $j^{\mathrm{th}}$ gene
    sequence of the genes belonging to the $i^{\mathrm{th}}$
    partition.
  \end{itemize}
  
  \noindent
  {\bf Class-Search method:} In eCGA, both the structure and the
  parameters of the model are searched and optimized to best fit the
  data. While the probabilities are learnt based on the variable
  instantiations in the population of selected individuals, a
  greedy-search heuristic is used to find an optimal or near-optimal
  probabilistic model. The search method starts by treating each
  decision variable as independent. The probabilistic model in this
  case is a vector of probabilities, representing the proportion of
  individuals among the selected individuals having a value '{\tt 1}'
  (or alternatively '{\tt 0}') for each variable. The model-search
  method continues by merging two partitions that yields greatest
  improvement in the model-metric score. The subset merges are
  continued until no more improvement in the metric value is possible.
  
\item {\em Offspring creation:\/} In eCGA, new individuals are created
  by sampling the probabilistic model. The offspring population are
  generated by randomly generating subsets from the current
  individuals according to the probabilities of the subsets as
  calculated in the probabilistic model.
  
\item {\em Replacement:\/} Many replacement schemes generally used in
  genetic and evolutionary computation---generational replacement,
  elitist replacement, niching, etc.---can be used in EDAs, but the
  key idea is to replace some or all the parents with some or all the
  offspring.
\item Repeat steps 2--6 until one or more termination criteria are met.
\end{enumerate}

\section{Sub-Structural Niching}
Traditional niching methods
\cite{Cavicchio:1970:phdThesis,DeJong:1975:phdThesis,Goldberg:1987:sharing,Mahfoud:1992:crowding,Yin:1993:niching,Harik:1995:RTS,Mahfoud:1995:phdThesis,Horn:1997:phdThesis,Mengshoel:1999:probabilisticCrowding}
achieve speciation by maintaining diversity at the level of
individuals. Effectiveness of such methods are strongly dependent on
the niche distribution. While some methods exist that can
automatically adjust the niche radius
\cite{Goldberg:1998:coevolutionarySharing}, they still detect
diversity on the individual level.\par

\begin{figure}
\center 
\subfigure[Trap: m = 10, k = 4]{\epsfig{file=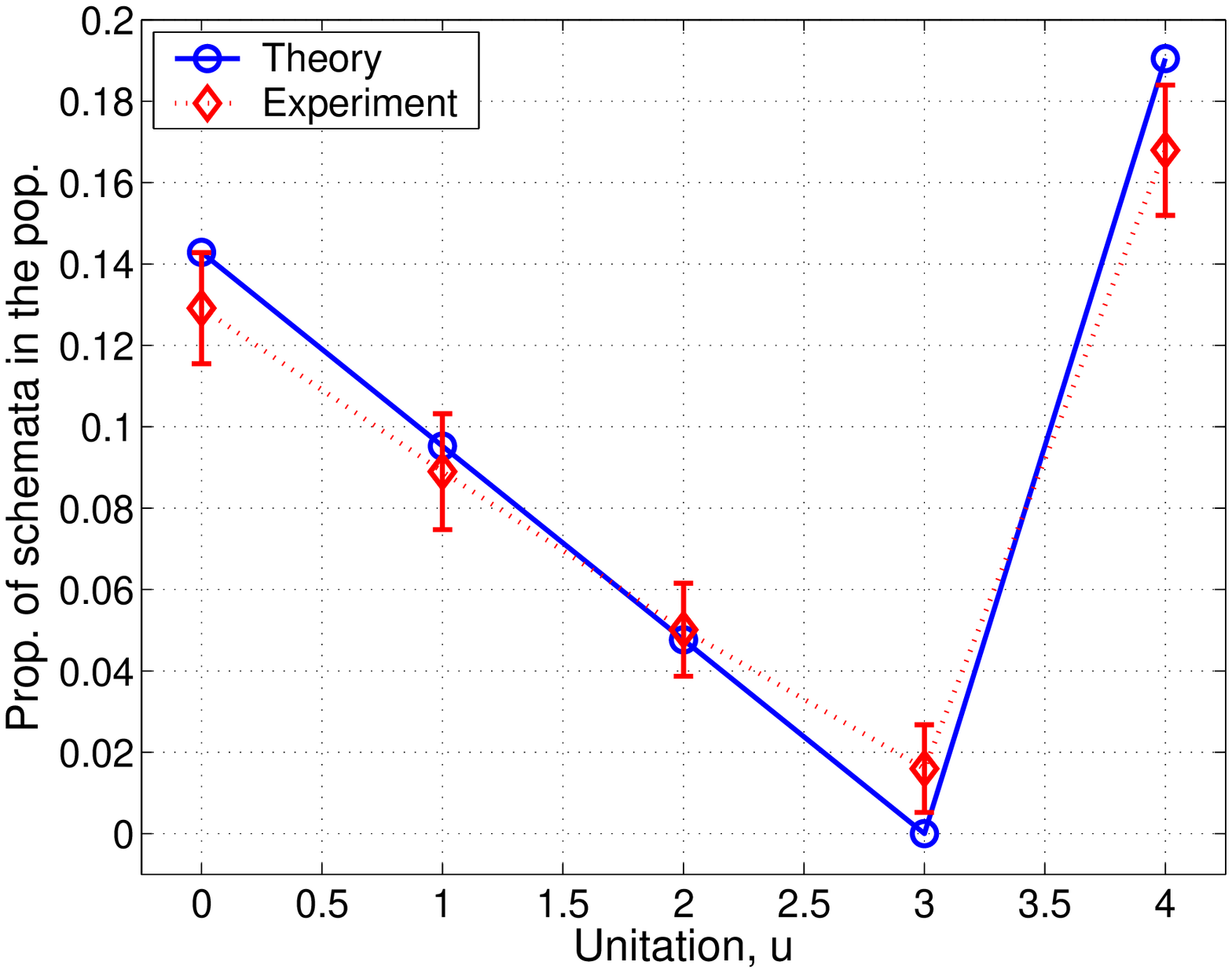,width=2.1in}}
\subfigure[Trap: m = 5, k = 5]{\epsfig{file=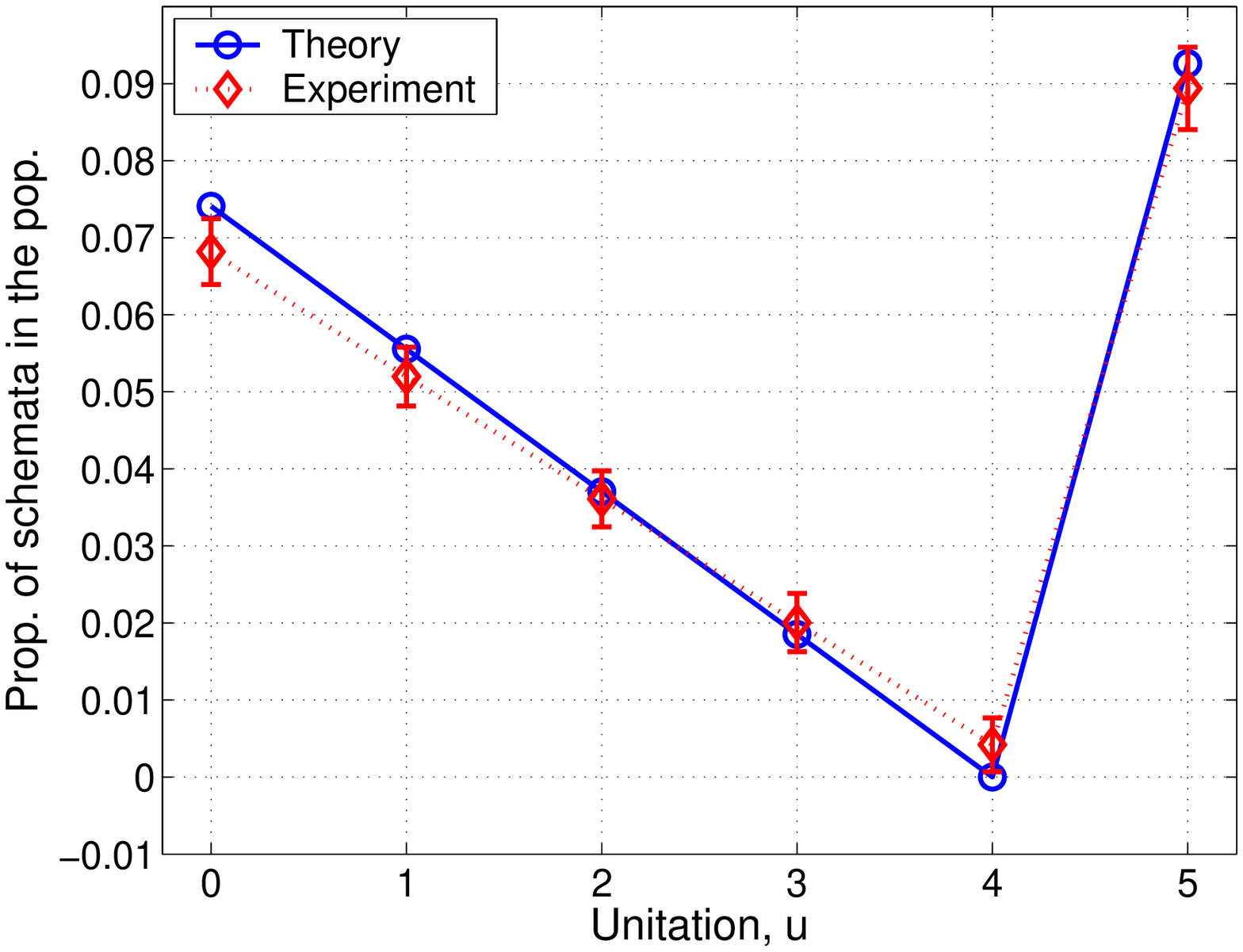,width=2.1in}}
\subfigure[Bipolar: m = 5, k = 6]{\epsfig{file=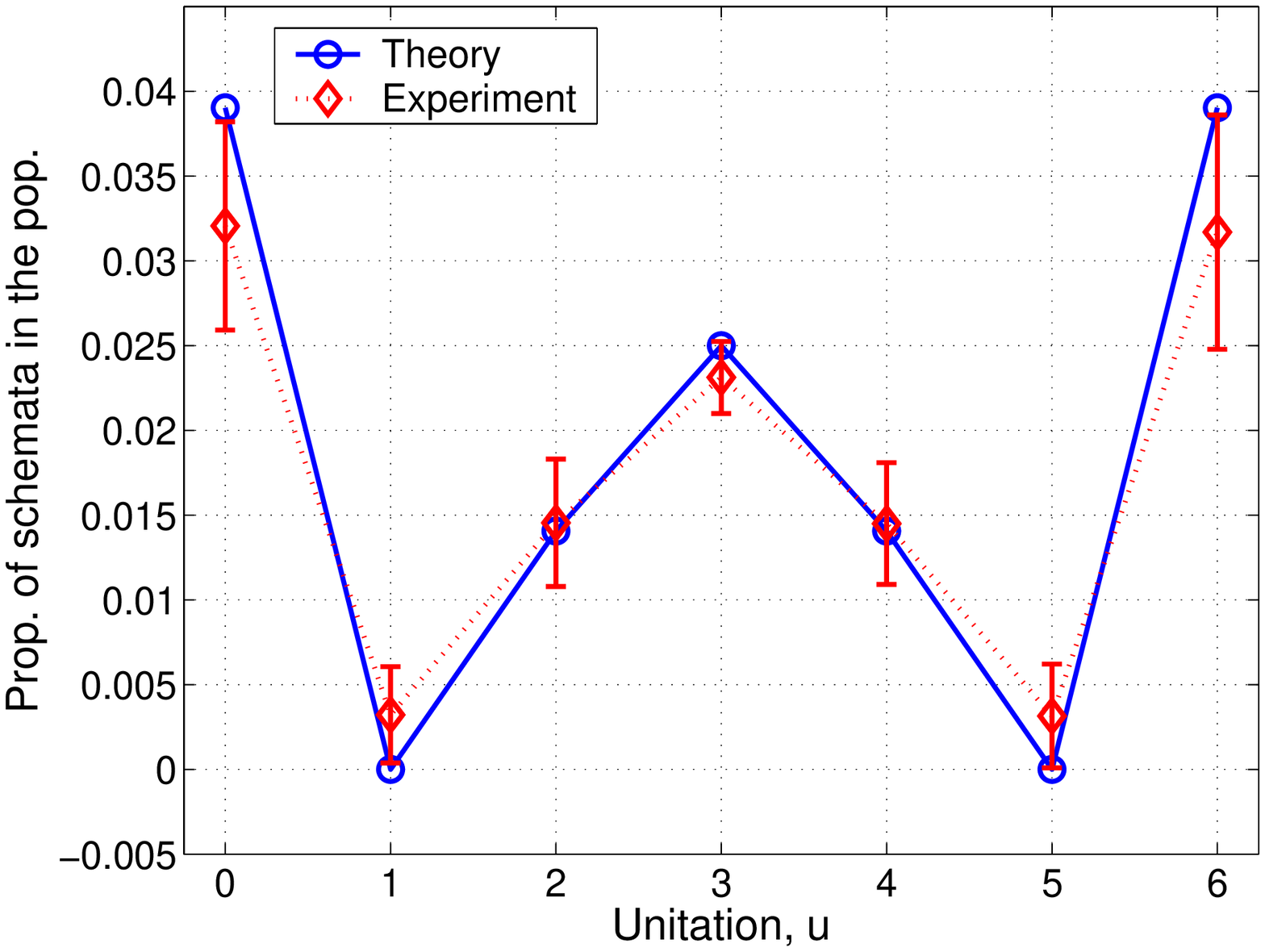,width=2.1in}}
\caption{Comparison of the ideal and experimental sub-structure
  frequencies for different additively separable problems.}
\label{fig:nichingVerification}
\end{figure}

One of the key elements of Goldberg's design decomposition
\cite{Goldberg:1991:designDecomposition,Goldberg:2002:DOI}---which has
been influential in the design and development of many competent
GAs---suggests that one of the critical steps for GA success is
problem decomposition, and identification and mixing of building
blocks. Since the EDAs work by first decomposing the search problems
into sub-structures and then creating new solutions by exchanging
different sub-structures, it might be advantageous, sometimes even
necessary, to maintain diversity at the building-block
(sub-structural) level and not at individual level
\cite{Sastry:2004:BBwiseNiching}. This is especially the case for
dynamic optimization, hierarchical-problem optimization, and
multiobjective optimization.\par

Sub-structural niching requires three key elements:\par
\noindent
{\bf Sub-structure identification:} To maintain diversity at
  the sub-structural level, we first need a mechanism to
  automatically identify all the important building blocks of the
  underlying search problem. In this study, we use the probabilistic
  models built by the eCGA. However, other linkage identification
  techniques
  \cite{Goldberg:1989:messyGA,Goldberg:1993:fastmGA,Kargupta:1996:gemGA,Munetomo:1999:LINC,Yu:2003:DSMGA,Harik:1997:phdThesis,Pelikan:2002:EDAsurvey,Larranaga:2002:EDAbook}
  can be used in a straightforward manner.\par
  
{\bf Sub-structure fitness estimation:} Once key sub-structures
  are identified, we must decide on which sub-structures to preserve
  and in what proportion. While sometimes, we might require to retain
  all sub-structure alternatives, usually we only need to preserve the
  highly fit ones. In order to do so, we need a way to estimate the
  quality of substructures from the fitnesses of individuals that
  possess them.\par In this study, we use the fitness-estimation
  method proposed by Sastry, Pelikan and Goldberg
  \cite{Sastry:2004:eCGAinheritance}. That is, after the probabilistic
  model is built and the linkage map is obtained, we estimate the
  fitness of sub-structures. In all, we estimate the fitness of a
  total of $\sum_{i = 1}^{m}2^{k_i}$ schemas. For example, for a
  four-bit problem, whose model is {\tt [1,3][2][4]}, the schemas
  whose fitnesses are estimated are: \{{\tt 0*0*}, {\tt 0*1*}, {\tt
    1*0*}, {\tt 1*1*}, {\tt *0**}, {\tt *1**}, {\tt ***0}, {\tt
    ***1}\}.\par
  
  The fitness of a sub-structure, $h$, is defined as the difference between
  the average fitness of individuals that contain the schema and the
  average fitness of all the individuals. That is,
\begin{equation}
\hat{f}_s(h) = {1\over n_h}\sum_{\{i|x_i \supset h\}}f\left(x_i\right) - {1 \over n{^\prime}}\sum_{i = 1}^{n^{\prime}}f\left(x_i\right)
\end{equation}
where $n_h$ is the total number of individuals that contain the schema
$h$, $x_i$ is the i$^{\mathrm{th}}$ individual and $f(x_i)$ is its
fitness, $n^{\prime}$ is the total number of individuals that were
evaluated. If a particular schema is not present in the population,
its fitness is arbitrarily set to zero. Furthermore, it should be
noted that the above definition of schema fitness is not unique and
other estimates can be used. The key point however is the use of the
probabilistic model in determining the schema fitnesses. Further
details regarding the estimation method are given elsewhere
\cite{Sastry:2004:eCGAinheritance,Pelikan:2004:BOAinheritance}.\par
 
\noindent
{\bf Sub-structure niche preservation:} Having identified the
  sub-structures and estimated their quality is not enough, we still
  need to decide on a methodology for preserving the substructures.
  Different methods such as fitness-proportionate, ranking, and
  truncation can be used and no one method is better than the other.
  For example, we can opt to preserve the sub-structures in proportion
  to their estimated fitness (so called fitness-proportionate method).
  That is, we have to modify the sampling frequencies of the
  sub-structures:
  \begin{equation}
    p_s\left(h_j\right) = {\hat{f}_s(h_j)\over\sum_{i = 1}^{2^k}\hat{f}_s(h_i)}
  \end{equation}
  We then use the above frequencies to sample the substructures to
  create new offspring.\par
  
  Regardless of how the sub-structure preservation is done, the key idea is to
  preserve those sub-structures that are potentially highly fit and
  are a part of different global optima.  The different sub-structure
  preservation methods usually requires modification of the sampling
  frequencies of sub-structures used in EDAs to generate new candidate
  solutions.\par
  
\begin{figure}
\center 
\epsfig{file=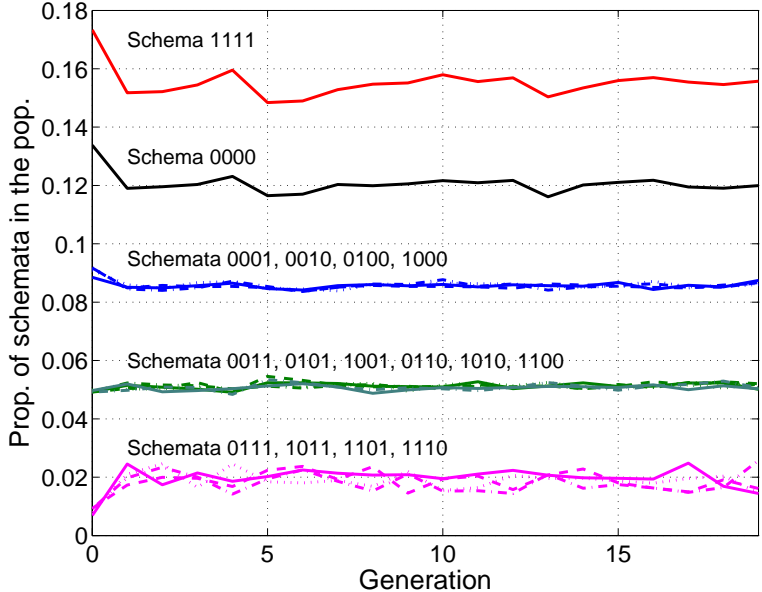,width=4in}
\caption{Illustration of sub-structure preservation via
  fitness-proportionate method for a 10-4 deceptive trap function}
\label{fig:nichingVerification1}
\end{figure}

  Before we use the proposed method for sustained maintenance of
  multiple global optima, we need to verify whether the sub-structure
  fitness estimate is accurate and if the method is capable of
  preserving different substructures over time. For the verification,
  we use fitness-proportionate method; that is, we maintain different
  substructures in proportion to their estimated fitness. We first
  investigate the accuracy of relative fitness estimates of different
  sub-structures in a given partition. The results for two different
  additively decomposable problems, m-k deceptive trap
  \cite{Ackley:1987:deception,Goldberg:1987:deception,Deb:1992:deception}
  and m-k bipolar function \cite{Goldberg:1992:deception} are shown in
  figure~\ref{fig:nichingVerification}. To demonstrate that the
  niching method can maintain all substructures in a sustained manner
  over time, we plot the market share of each of the 16 schemata for
  the 10-4 deceptive trap functions in
  figure~\ref{fig:nichingVerification1}. The results show that the
  substructure-fitness estimation is quite accurate and that it can
  preserve the substructures at their desired proportions over time.

\section{Results and Discussion}\label{sec:results}
In this section, we investigate the effectiveness of the
sub-structural niching in stably maintaining all the global optima
over a large number of generations, and the population size required to
do so, as a function of the number of optima. We note that
in all the results presented in this paper, we consider
fitness-proportionate sub-structural niche preservation mechanism for
proof-in-principle and, as mentioned earlier, other possible mechanisms
can be more beneficial depending on the goal for using the niching
mechanism. Additionally, the window size for RTS was set to the
problem size, as suggested elsewhere \cite{Pelikan:2001:hBOA}. Before
presenting the results we first give a brief description of the test
problem considered in the experiments.\par

Our approach in verifying the performance of sub-structural niching is
to consider bounding {\em adversarial problems\/} that exploit one or
more dimensions of problem difficulty \cite{Goldberg:2002:DOI}.
Particularly, we are interested in problems where building-block
identification is critical for the GA success. Additionally, the
problem solver (eCGA) should not have any knowledge of the
building-block structure of the test problem, but should be known to
researchers for verification purposes.\par

One such class of problems is the m-k deceptive {\em trap\/} problem,
which consists of additively separable {\em deceptive\/} functions
\cite{Ackley:1987:deception,Goldberg:1987:deception,Deb:1992:deception}.
Deceptive functions are designed to thwart the very mechanism of
selectorecombinative search by punishing any localized hillclimbing
and requiring mixing of whole building blocks at or above the order of
deception. Using such {\em adversarially\/} designed functions is a
stiff test---in some sense the stiffest test---of algorithm
performance. The idea is that if an algorithm can beat an
adversarially designed test function, it can solve other problems that
are equally hard or easier than the adversarial function.\par

 In this study, we use a modified m-4 deceptive trap problems where both 0000
and 1111 have equal fitness. Therefore there are $2^m$ global optima
with an identical fitnesses. That is, each $k$-bit trap is
defined as follows:
\begin{equation}
trap_k(u) = 
\left\{
\begin{array}{ll}
1 & \mbox{~~if $u=k$} \\
1 & \mbox{~~if $u=0$} \\
0.75\left[1-{u\over k-1}\right] & \mbox{~~otherwise}
\end{array}
\right.,
\end{equation}
where $u$ is the number of $1$s in the input string of $k$ bits.\par

\begin{figure}
\center 
\subfigure[RTS]{\label{fig:RTSsingleRun}\epsfig{file=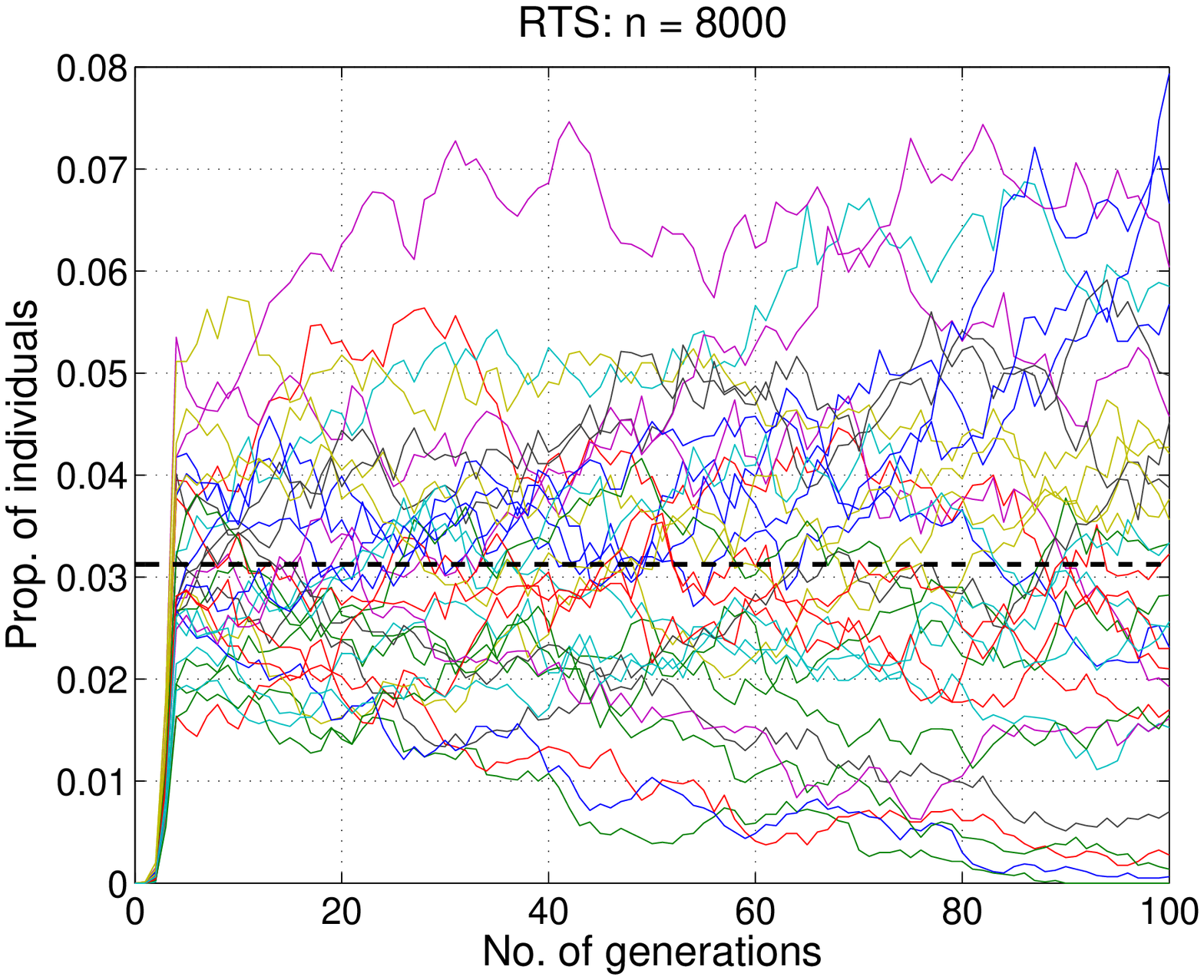,width=2.8in}}
\subfigure[Sub-structural niching]{\label{fig:BBWsingleRun}\epsfig{file=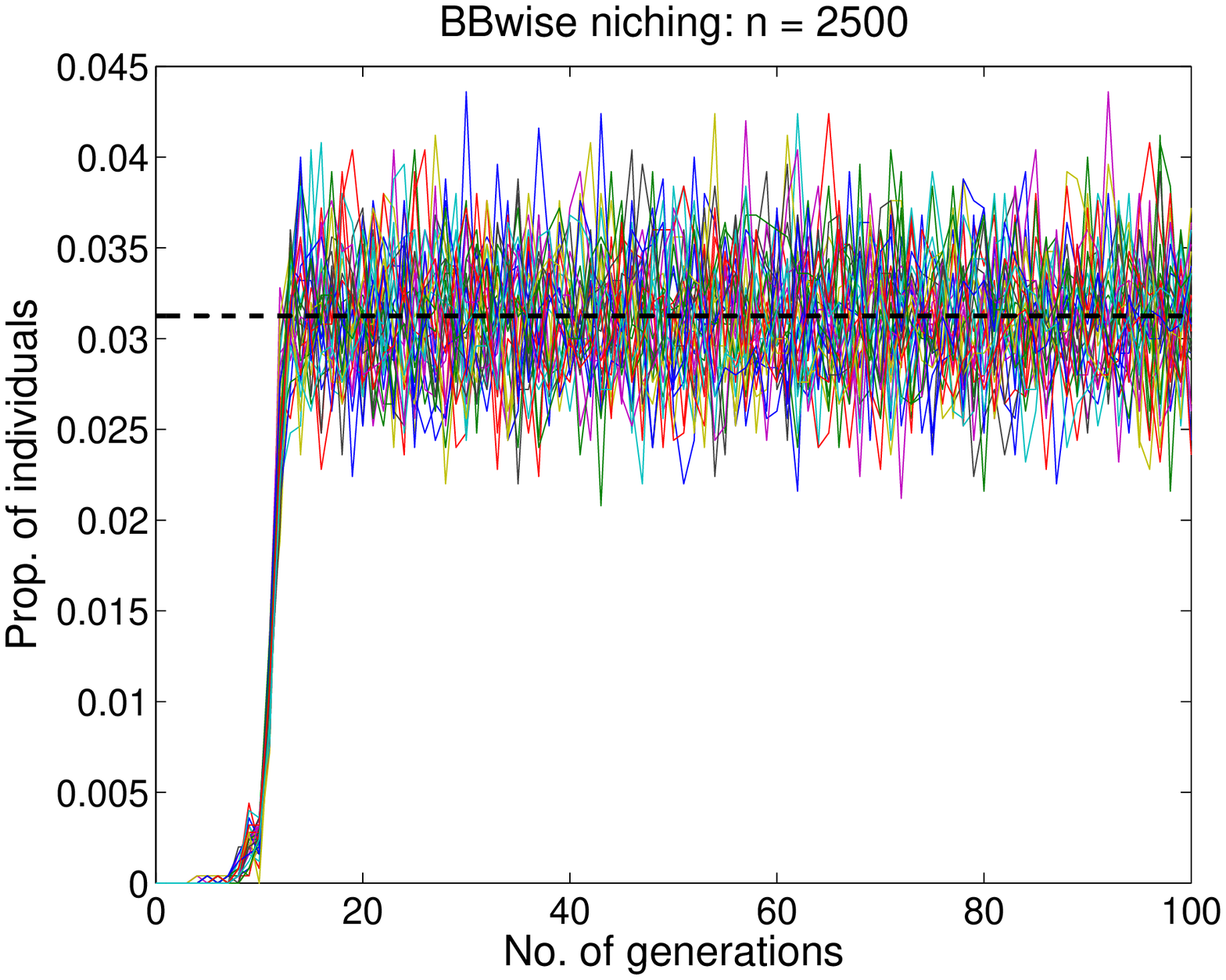,width=2.8in}}
\subfigure[RTS]{\label{fig:RTSavg}\epsfig{file=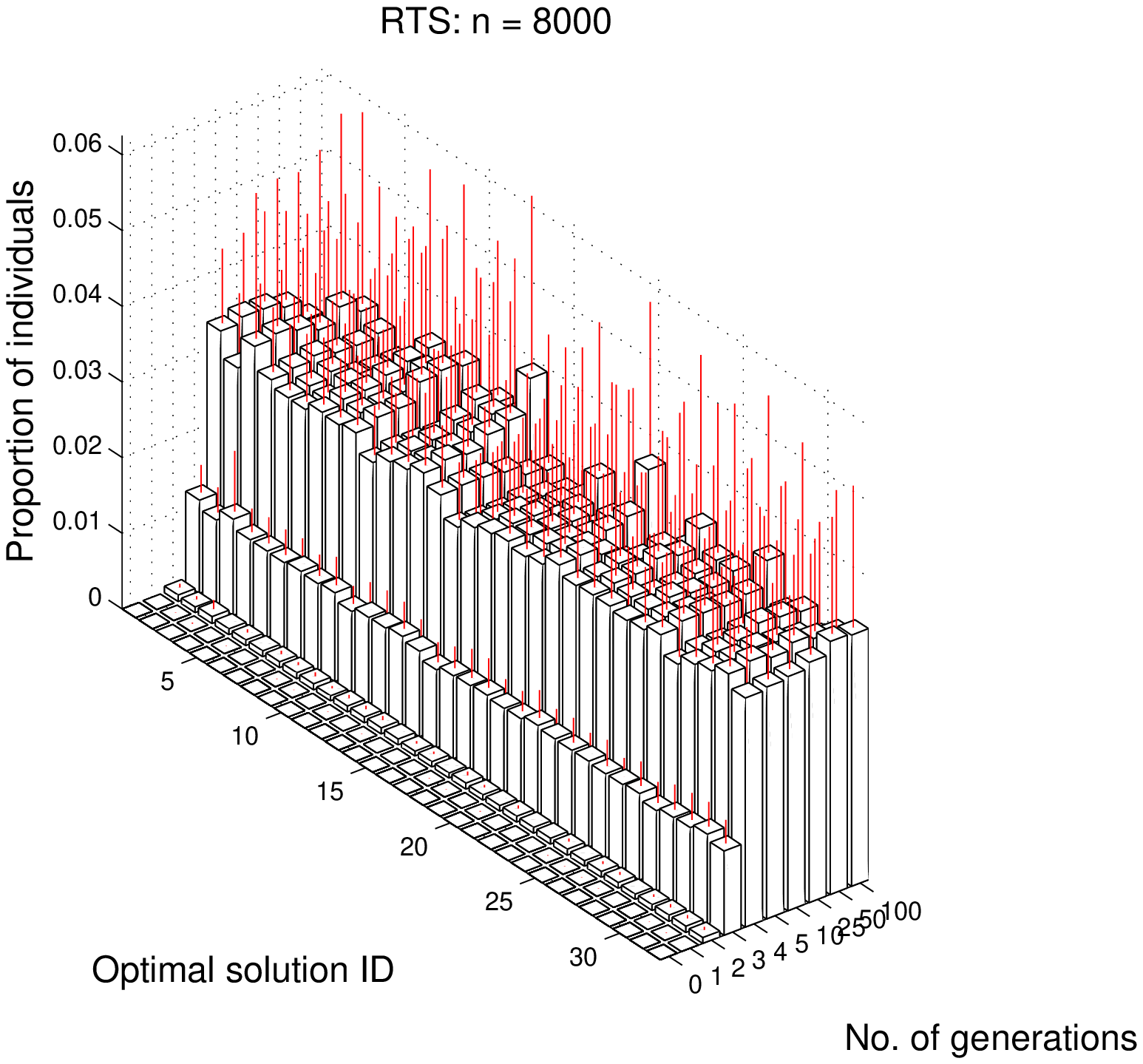,width=2.8in}}
\subfigure[Sub-structural niching]{\label{fig:BBWavg}\epsfig{file=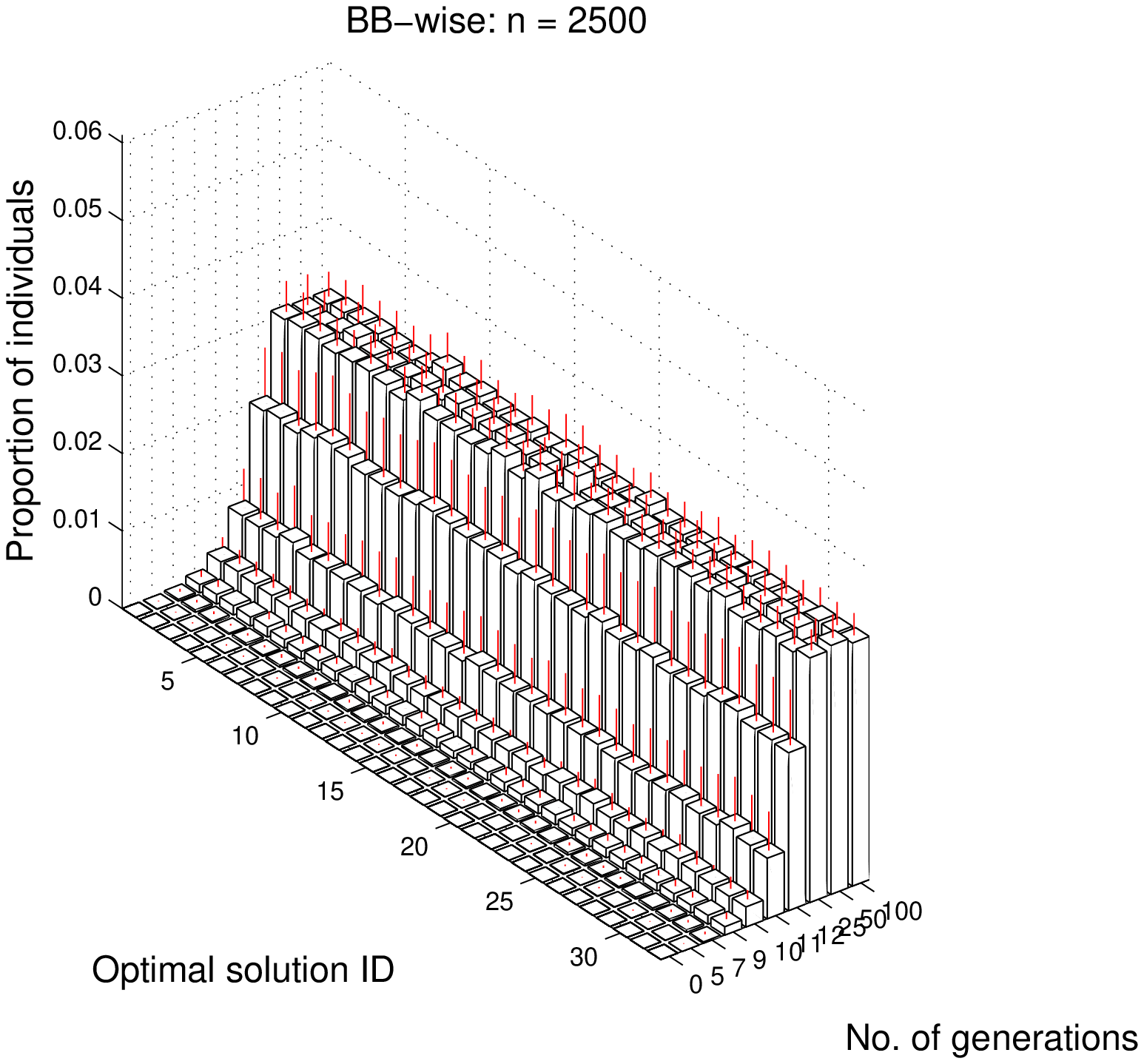,width=2.8in}}
\subfigure[RTS]{\label{fig:RTSavgAvg}\epsfig{file=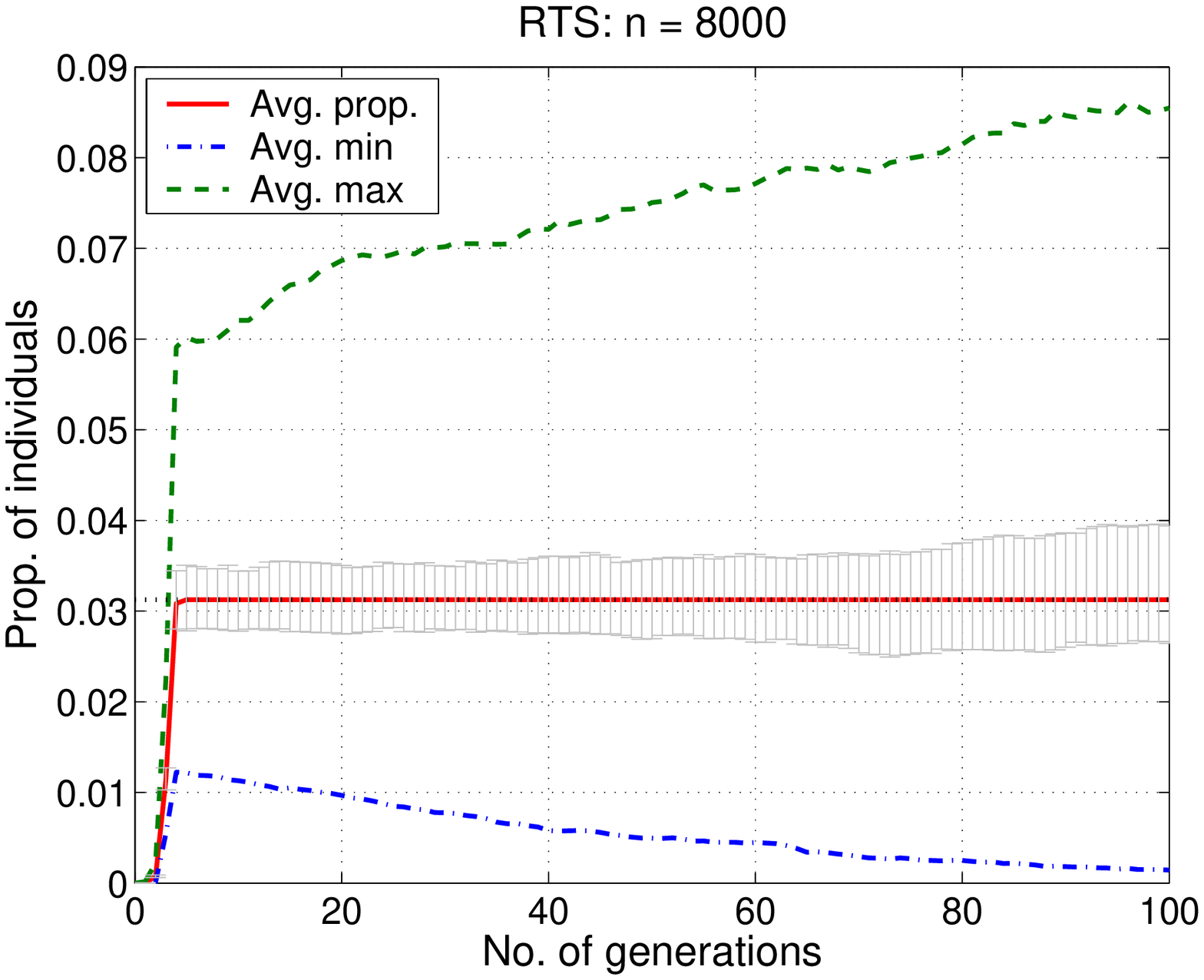,width=2.8in}}
\subfigure[Sub-structural niching]{\label{fig:BBWavgAvg}\epsfig{file=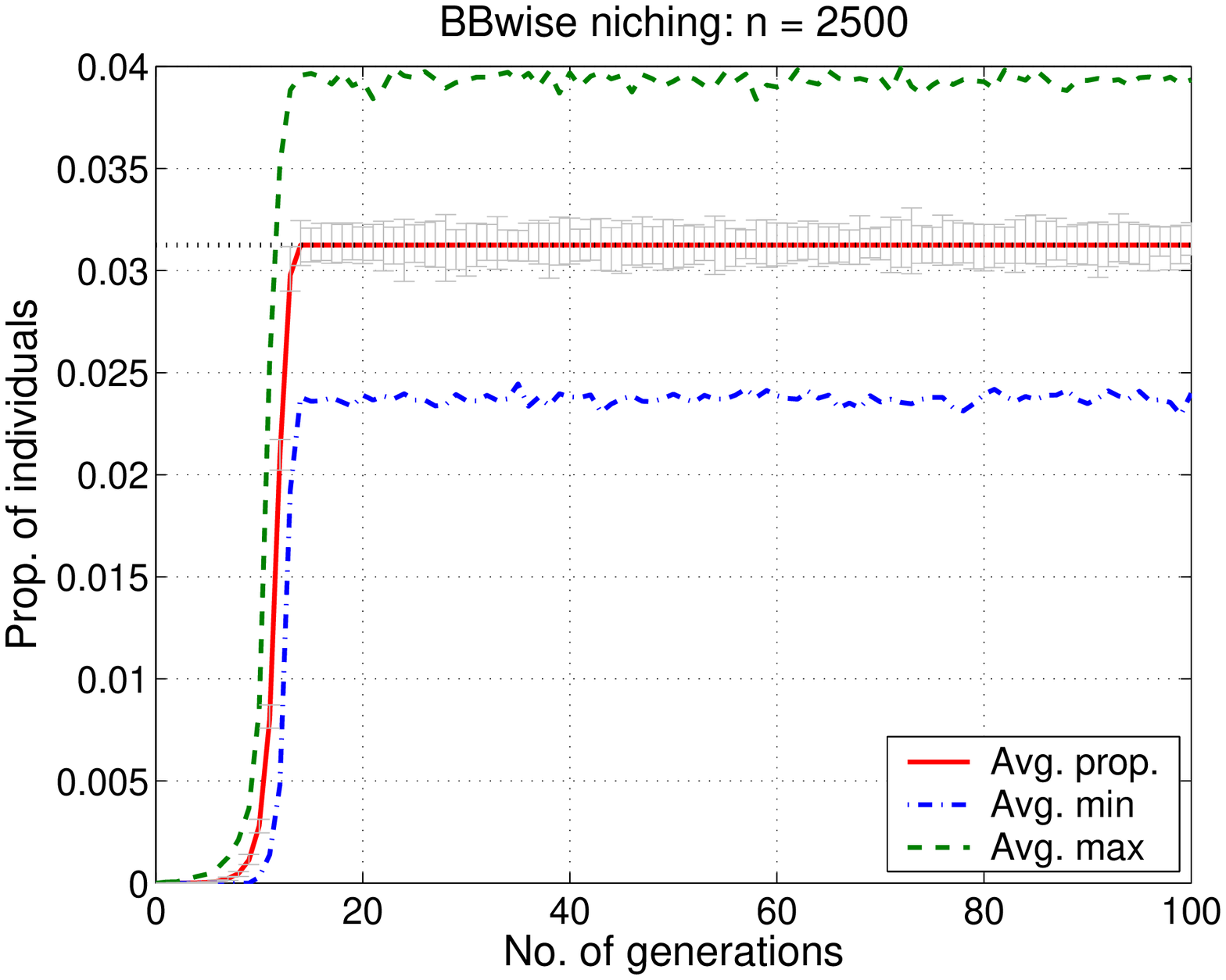,width=2.8in}}
\caption{Comparison of the performance of RTS and sub-structural
  niching in stably maintaining all global optima for a concatenated
  5-4 deceptive trap problem: (a) \& (b) Single GA run behavior, (c)
  \& (d) Average behavior, and (e) \& (f) Average, average minimum and
  average maximum proportion allocated to an optima. The maintenance
  of all the optima by RTS is very noisy and unstable, while,
  sub-structural niching maintains all the niches stably over large
  number of generations. Additionally, the market share of each optima
  in sub-structural niching is close to the expected proportion of
  0.03125. Results in (c)-(f) are averaged over 50 independent GA
  runs.}
\label{fig:RTSvsBBW}
\end{figure}

First, we compare the ability of RTS and sub-structural niching in
maintaining all the global optima over time in a stable manner (see
figure~\ref{fig:RTSvsBBW}). We start by comparing the single GA run
behavior of both niching methods in figures~\ref{fig:RTSsingleRun} and
\ref{fig:BBWsingleRun}, where we show the proportion of individuals in
each of the 32 global optima of a 5-4 trap function as a function of
time. The results clearly show that in contrast to RTS the niche
maintenance of sub-structural niching is highly stable and the allocated
market share for each optima is in agreement with the desired
proportion of $1/32 = 0.03125$\footnote{Since all the global optima
  have identical fitness, we expect that the market share of each
  optima is 1/32 = 0.03125.}. We then consider the average behavior of
both RTS and sub-structural niching, where we plot the average market
share of each of the global optima over time in
figures~\ref{fig:RTSavg} and \ref{fig:BBWavg}. The lines above the
bars in figures~\ref{fig:RTSavg} and \ref{fig:BBWavg} depict the
standard deviation and optimal solution ID refers to an arbitrary (but
unique) number for each of the 32 global optima. For simplicity, we
also plot the average, average minimum, and average maximum market
share of an optima in figures~\ref{fig:RTSavgAvg} and
\ref{fig:BBWavgAvg}. Figures~\ref{fig:RTSavg}--\ref{fig:BBWavgAvg}
clearly show that RTS cannot stably maintain the global optima, even
at large population sizes, when compared to sub-structural
niching.\par

Figure~\ref{fig:RTSvsBBW} clearly indicates overall the effectiveness
of sub-structural niching in stably maintaining all the global optima
at the desired proportion over large number of generations. We note
that the time to detect the global optima is faster in RTS than in
sub-structural niching. This is to be expected as sub-structural
niching maintains diversity in all sub-structures proportional to
their fitness, and it takes longer for mixing to hone in on to the
global optima. However, once the optima are found, sub-structural
niching preserves much more stably than RTS.\par

\begin{figure}
\center
\subfigure[RTS]{\label{fig:rtsQvsN} \epsfig{file=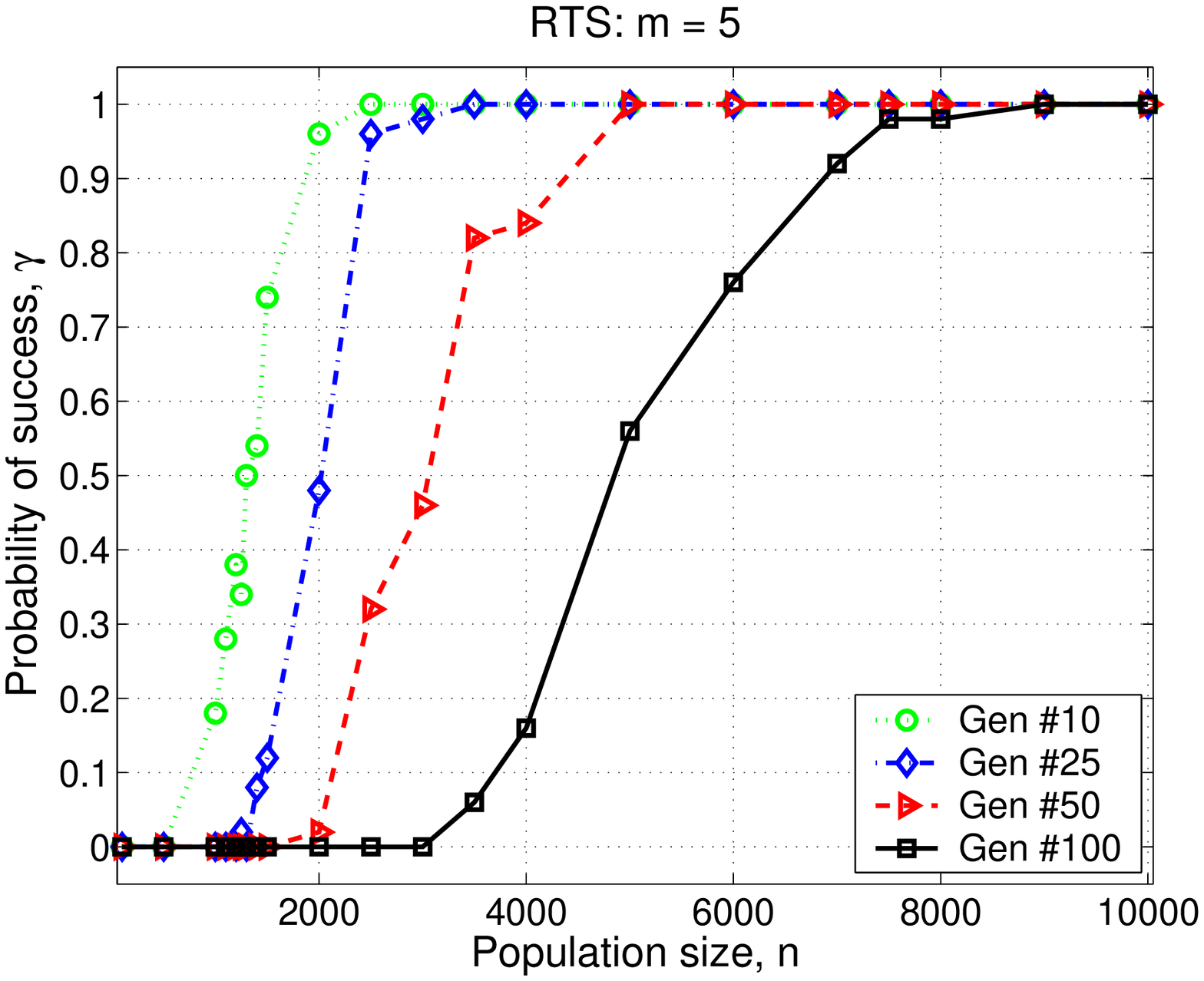,width=3in}}
\subfigure[Sub-structural niching]{\label{fig:bbwQvsN} \epsfig{file=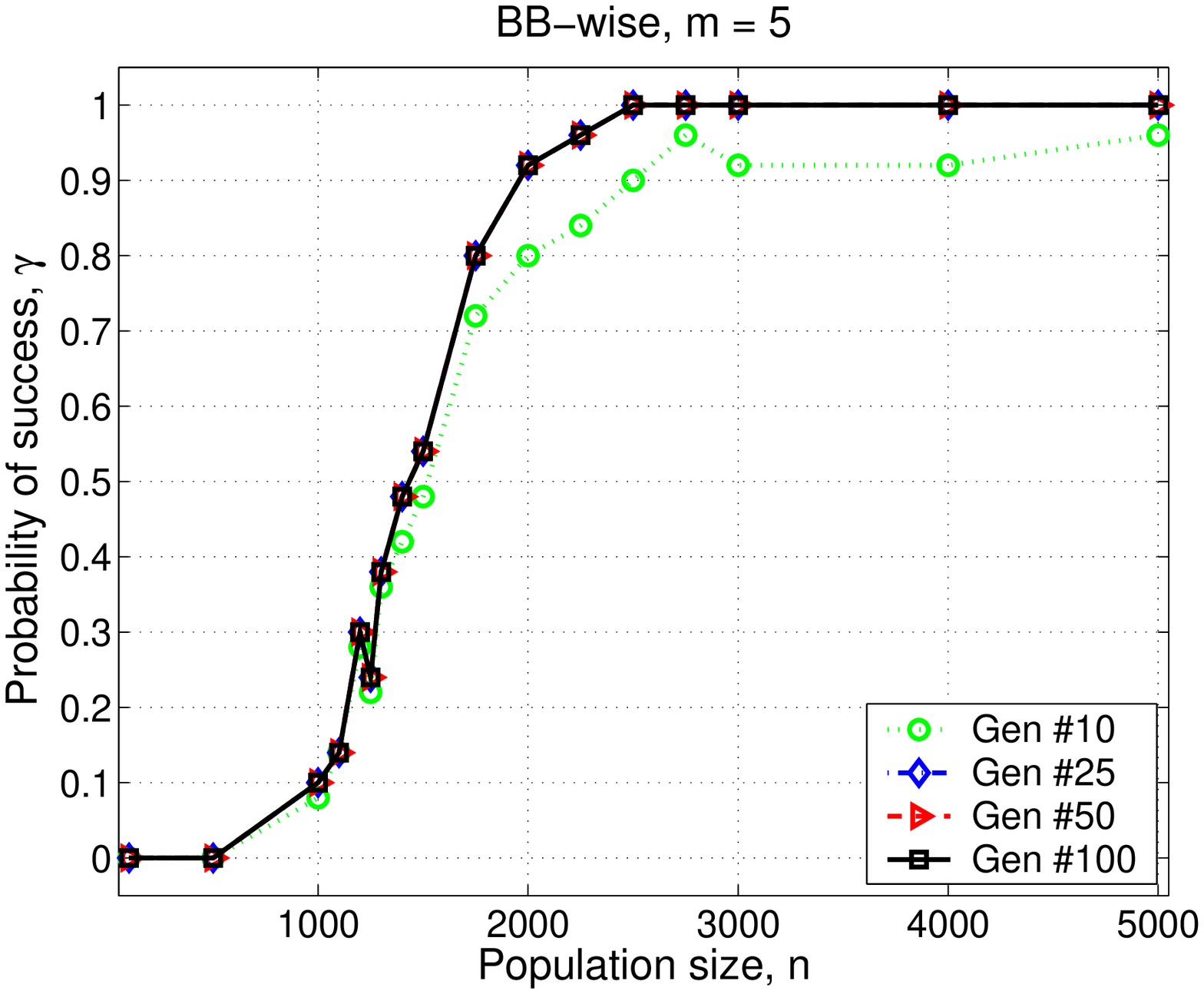,width=3in}}
\caption{The probability of maintaining at least one copy of all the
  global optima, $\gamma$, over different number of generations as a
  function of population size for RTS and sub-structural niching. RTS
  requires significantly larger population size to maintain all the
  global optima than the sub-structural niching. The results are
  averaged over 50 independent GA runs}
\label{fig:RTSvsBBWn}
\end{figure}

We also studied the effect of population size, $n$, on the success
probability of maintaining the global optima, $\gamma$, the results of
which are shown for 5-4 deceptive trap function in
figure~\ref{fig:RTSvsBBWn}. The figure plots the probability of
maintaining all the global optima for different number of generations
as a function of population size for both RTS and sub-structural
niching. As shown in the figure, RTS, requires larger population sizes
to maintain the global optima for longer time. This is well understood
phenomena of traditional nichers and has been analyzed by Mahfoud
for fitness sharing \cite{Mahfoud:1994:nichingPopSizing}. However, in
sub-structural niching, the population size required to achieve a
certain success probability, $\gamma$, is independent of the number of
generations we would like to maintain the niches. Additionally, RTS
requires significantly larger population size than sub-structural
niching to achieve the same level of success probability.\par

\begin{figure}
\center
\epsfig{file=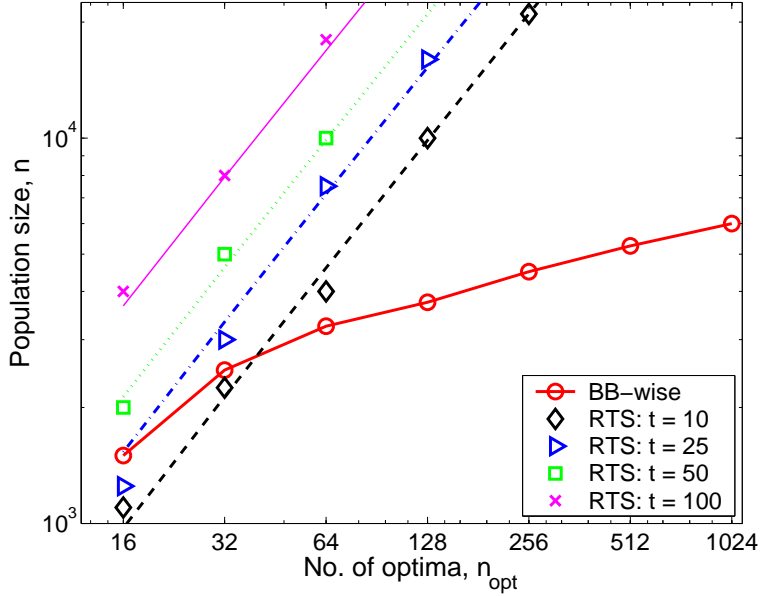,width=4in}
\caption{Minimum population sizing required for maintaining at least
  one copy of $n_{opt}-1$ optima ($\gamma = (n_{opt}-1)/n_{opt} =
  (2^m-1)/2^m$) over different number of generations as a function of
  number of optima for RTS and sub-structural niching. The results for
  RTS conform to the population-sizing model of Mahfoud [23]
  (equation~\ref{eqn:popSizing}). The results are averaged over 50
  independent GA runs.}
\label{fig:popSizing}
\end{figure}

Finally, we use the $n$ versus $\gamma$ results, to determine the
population size required to maintain successfully all the global
optima with high probability. Specifically, we plot the minimum
population size required by sub-structural niching and RTS for
maintaining at least one copy of $n_{opt}-1$ or more global optima in
the population for different number of generations as a function of
number of optima, $n_{opt}$, is figure~\ref{fig:popSizing}. The lines
plotted for the RTS results are from the population-sizing model of
Mahfoud \cite{Mahfoud:1994:nichingPopSizing}:
\begin{equation}
\label{eqn:popSizing} n \propto {\log\left[\left(1 -
    \gamma^{1/t}\right)/n_{opt}\right] \over \log\left[\left(n_{opt} - 1\right)/n_{opt}\right]}
\end{equation}
where $t$ is the number of generations we need to maintain all the
niches. Figure~\ref{fig:popSizing} clearly shows that sub-structural
niching requires significantly less population size than RTS to
maintain all the global solutions with a high probability. We note
that the population size for sub-structural niching eventually will
grow linearly (in accordance with the population-sizing model) with
n$_{opt}$ as we need at least one individual for each of the optima.
However, this might not be the case for hierarchical and dynamic
optimization problems, where diversity is required only at the
sub-structural level and not at the solution level. Nevertheless,
sub-structural niching requires orders of magnitude smaller populations
than RTS and stably maintains niches with a high probability even for
a problem with about a thousand global optima.

\section{Summary and Conclusions}
In this paper we proposed a sub-structural niching mechanism, which, in
contrast to traditional niching mechanisms, exploits the problem
decomposition capability of estimation of distribution algorithms and
stably maintains diversity at the sub-structure, or building-block
level rather than at individual level. The sub-structural niching
mechanism consists of three components: (1) Sub-structure
identification, where we use the probabilistic model built by EDAs,
specifically, extended compact GA \cite{Harik:1999:eCGA}, (2)
sub-structure fitness estimation, where we use the fitness-estimation
procedure proposed by Sastry {\em et al\/}
\cite{Sastry:2004:eCGAinheritance}, and (3) sub-structure niche
preservation, where different mechanisms are can be envisioned, each
suitable based on the purpose and objective of using the niching
method. Regardless of how it is done, the key idea of the sub-structure niche
preservation mechanism is to preserve highly-fit sub-structures in
desired proportions in the population in a stable manner over the
duration of the search.\par

We also tested performance of the sub-structural niching mechanism on a
class of boundedly-difficult additively separable multimodal problems
and compared it with those of restricted tournament selection
(RTS)---a niching method used in hierarchical Bayesian optimization
algorithm. The results show that not only is the sub-structural
niching mechanism able to stably preserve multiple global optima over
large number of generations, but does so with a high probability
requiring significantly less population size than RTS. The results indicate
that sub-structural niching can be particularly effective with
hierarchical and dynamic problem optimization.

\section*{Acknowledgments}
This work was sponsored by the Air Force Office of Scientific
Research, Air Force Materiel Command, USAF, under grant
F49620-03-1-0129, the National Science Foundation under ITR grant
DMR-03-25939 at Materials Computation Center, and ITR grant
DMR-01-21695 at CPSD, and the Dept. of Energy under grant
DEFG02-91ER45439 at Fredrick Seitz Materials Research Lab. The U.S.
Government is authorized to reproduce and distribute reprints for
government purposes notwithstanding any copyright notation thereon.

\bibliographystyle{my-apa-uiuc}
\bibliography{kumaraBibliography}
\end{document}